\documentclass[10pt,twocolumn,letterpaper]{article}

\usepackage[pagenumbers]{wacv} 

\usepackage{graphicx}
\usepackage{amsmath}
\usepackage{amssymb}
\usepackage{booktabs}

\usepackage{graphicx}
\usepackage[table,xcdraw,dvipsnames]{xcolor}
\usepackage{multirow}

\usepackage{enumitem}
\newcommand{\argmin}{\mathop{\mathrm{argmin}}}

\setlist{nosep}

\usepackage[pagebackref,breaklinks,colorlinks]{hyperref}

\usepackage[capitalize]{cleveref}
\crefname{section}{Sec.}{Secs.}
\Crefname{section}{Section}{Sections}
\Crefname{table}{Table}{Tables}
\crefname{table}{Tab.}{Tabs.}

\def\wacvPaperID{749} 
\def\confName{WACV}
\def\confYear{2024}

\begin{document}

\title{ReConPatch : Contrastive Patch Representation Learning for Industrial Anomaly Detection}

\author{Jeeho Hyun
\qquad
Sangyun Kim
\qquad
Giyoung Jeon
\qquad
Seung Hwan Kim
\and
Kyunghoon Bae
\qquad
Byung Jun Kang\thanks{Correspondence to: \texttt{bj.kang@lgresearch.ai}}\\
\\
LG AI Research
}
\maketitle

\begin{abstract}
Anomaly detection is crucial to the advanced identification of product defects such as incorrect parts, misaligned components, and damages in industrial manufacturing.  Due to the rare observations and unknown types of defects, anomaly detection is considered to be challenging in machine learning.  To overcome this difficulty, recent approaches utilize the common visual representations pre-trained from natural image datasets and distill the relevant features. However, existing approaches still have the discrepancy between the pre-trained feature and the target data, or require the input augmentation which should be carefully designed, particularly for the industrial dataset. In this paper, we introduce ReConPatch, which constructs discriminative features for anomaly detection by training a linear modulation of patch features extracted from the pre-trained model. ReConPatch employs contrastive representation learning to collect and distribute features in a way that produces a target-oriented and easily separable representation. To address the absence of labeled pairs for the contrastive learning, we utilize two similarity measures between data representations, pairwise and contextual similarities, as pseudo-labels. Our method achieves the state-of-the-art anomaly detection performance (99.72\%) for the widely used and challenging MVTec AD dataset. Additionally, we achieved a state-of-the-art anomaly detection performance (95.8\%) for the BTAD dataset.
\end{abstract}

\section{Introduction}
Anomaly detection in industrial manufacturing is key to identify the defects in products and maintain their quality. Anomalies can include incorrect parts, misaligned components, or damage to the product. Machine learning approaches for anomaly detection have been widely studied with an increasing demand for the automation in industrial applications.
The main concern of these approaches is to learn how to discriminate anomalies from normal cases based on previously collected data. However, anomaly detection is particularly challenging because defects are rarely observed and unknown types of defects can occur. Such situation, in which the majority of cases are marked as normal and abnormal cases are scarce in the collected data, has lead to the improvements in one-class classification. 

The key concept of one-class classification for anomaly detection is to train a model to learn a distance metric between data and detect anomalies at a large distance from the nominal data. In an effort to learn the metric, reconstruction-based approaches have been proposed to detect anomalies by measuring the reconstruction errors using auto-encoding models \cite{davletshina2020unsupervised,nguyen2019anomaly,sakurada2014anomaly} or generative adversarial networks (GANs) \cite{pidhorskyi2018generative,sabokrou2018adversarially}. As the variety of data is not sufficiently rich to estimate a reliable nominal distribution from scratch, recent works have shown that leveraging the common visual representation, obtained from a natural image dataset \cite{deng2009imagenet}, can result in high anomaly detection performance \cite{bergman2020deep,cohen2020sub}. 
Although pre-trained models can provide rich representations without adaptation, such representations are not sufficiently distinguishable to identify subtle defects in industrial images. The distribution shift between natural and industrial images also makes it difficult to extract anomaly-specific features. 
For improvements in anomaly detection performance, it is essential to train a model to learn a representation space that effectively discriminates borderline anomalies.

To alleviate the distribution shift between the pre-trained and the industrial datasets, prominent features for anomaly detection can be distilled by training a student model to reproduce the representation of the pre-trained model using a teacher supervision \cite{bergmann2020uninformed}. Attaching a normalizing flow \cite{dinhdensity} at the end of the pre-trained model is another approach to exploit the pre-trained representation and estimate the distribution of normality \cite{rudolph2021same}. 
Unfortunately, existing methods still require extensive handcrafted input augmentation, such as random crop, random rotation, or color jitter.
Particularly in case of industrial images, data augmentation should be carefully designed by the user expertise.

In this paper, we introduce unsupervised metric learning framework for anomaly detection by enhancing the arrangement of the features, \textit{ReConPatch}. Contrastive learning-based training schemes present weaknesses in terms of modeling variations within nominal instances, which may increase the false-positive rate of the anomaly detection. To this end, ReConPatch utilizes the contextual similarity \cite{kim2022self} among features obtained from the model as a pseudo-label for the training.
Specifically, our method efficiently adapts feature representation by training only a simple linear transformation, as opposed to training the entire network.
By doing so, we are able to learn a target-oriented feature representation which achieves higher anomaly detection accuracy without input augmentation, making our method a practical and effective solution for anomaly detection in various industrial settings.

\begin{figure*}
\centering
\includegraphics[width=0.9\linewidth]{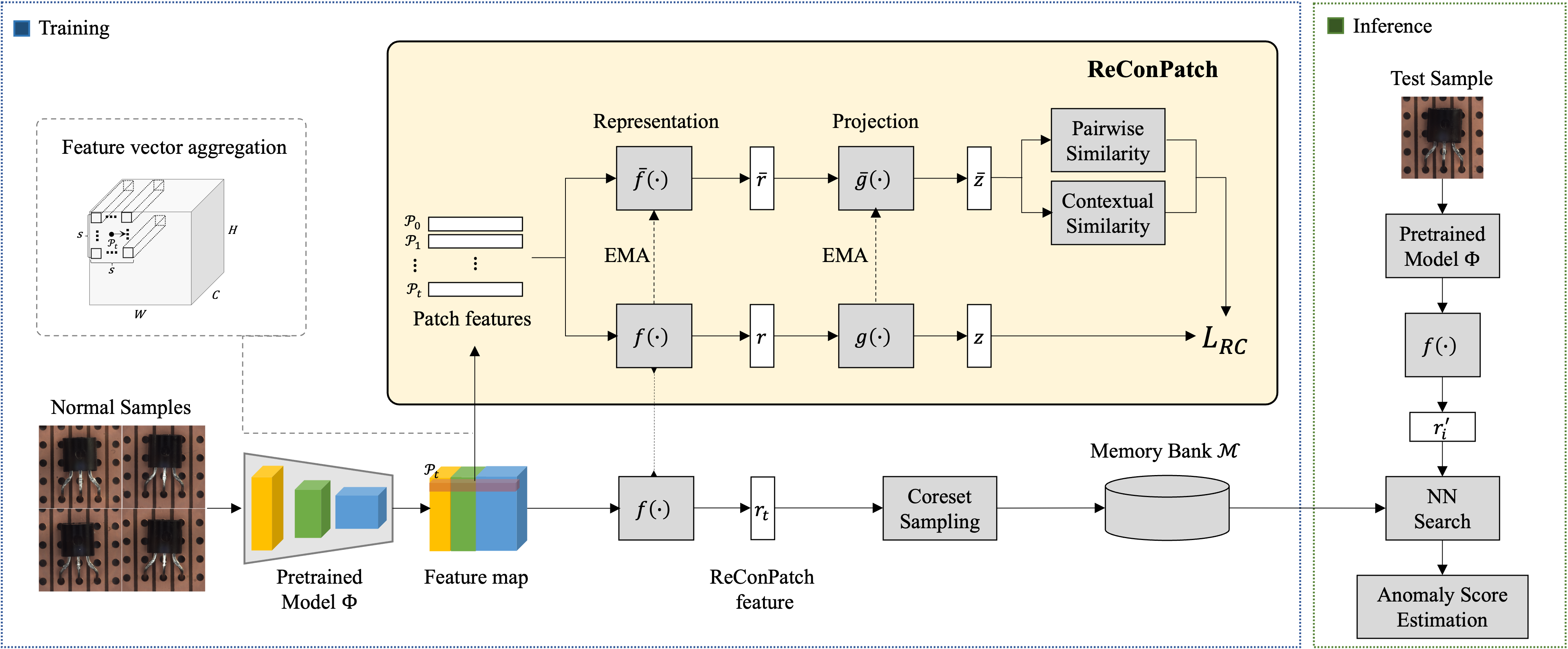}
\caption{Overall structure of the anomaly detection using ReConPatch. ReConPatch consists of two networks to train representations of the patch-level features, which includes the feature representation layer $f$, $\bar{f}$ and projection layer $g$, $\bar{g}$ respectively. Upper networks ($\bar{f}, \bar{g}$) are used to calculate pairwise and contextual similarities between patch-level feature pairs, while the bottom networks ($f, g$) used for the representation learning of patch-level features is trained through relaxed contrastive loss $\mathcal{L}_{RC}$.}
\label{fig:overall_structure}
\end{figure*}
\section{Related Work}
Unsupervised machine learning approaches in anomaly detection using neural networks have been widely analyzed. 
Deep Support Vector Data Description (SVDD) trains a neural network to map each datum to the hyperspherical embedding and detect anomalies by measuring the distance from the center of the hypersphere \cite{ruff2018deep}. 
Patch SVDD has been developed as a patch-wise extension of Deep SVDD, utilizing the features of each spatial patch from the convolutional neural network (CNN) feature map to enhance localization and enable fine-grained examination \cite{yi2020patch}.
The reconstruction-based approach assumes that normal data can be accurately reconstructed or generated by training a model using a nominal dataset, whereas abnormal data cannot. Based on this assumption, an anomaly score is calculated as the error between the original input and the reconstructed input.
Auto-encoding models are used for the reconstruction model \cite{davletshina2020unsupervised,nguyen2019anomaly,sakurada2014anomaly}. With the improvements in GANs, several approaches have also shown the effectiveness of GANs in anomaly detection \cite{pidhorskyi2018generative,sabokrou2018adversarially}. When training a model from scratch, variety and abundance should be guaranteed, which is mostly not available for anomaly detection.

To alleviate the shortage of data in anomaly detection, several attempts have been made to utilize a common visual representation pre-trained with a rich natural image dataset \cite{deng2009imagenet}.
Previous studies that use such pre-trained model measures the distance between the representations of input data and their nearest neighbors to detect the anomalies \cite{bergman2020deep} and compares hierarchical sub-image features to localize anomalies \cite{cohen2020sub}. To efficiently compare the input with training set, a memory bank is introduced to store the representatives \cite{cohen2020sub}.

DifferNet \cite{rudolph2021same} provides a normalizing flow \cite{dinhdensity} that is helpful in training a bijective mapping between the pre-trained feature distribution and the well-defined density of the nominal data, which is used to identify the anomalies.
A condition normalizing flow using positional encoding is proposed by CFLOW-AD \cite{gudovskiy2022cflow}. As the normalizing flow is trained to map features to the nominal distribution, this method is vulnerable to the outliers in the training dataset.

PatchCore proposes a locally aware patch feature and efficient greedy subsampling method to define the coreset \cite{roth2022towards}.
The coupled-hypersphere-based feature adaptation (CFA) trains a patch descriptor that maps features onto the hypersphere, which is centered on the nearest neighbor in the memory bank \cite{lee2022cfa}.
PaDiM estimates a Gaussian distribution of patch features at each spatial location to detect and localize out-of-distributions (OODs) as anomalies \cite{defard2021padim}.
PNI is developed to train a neural network to predict the feature distribution of each spatial location and its neighborhoods \cite{bae2023pni}. 

\section{Method}
Our proposed method, ReConPatch, focuses on learning a representation space that maps features extracted from nominal image patches to be grouped closely if they share similar nominal characteristics in an unsupervised learning manner. 
Although previous work \cite{roth2022towards} has shown the effectiveness of selecting representative nominal patch features using a pre-trained model, this model still presents a representation biased to the natural image data, which has a gap with the target data.
The main concept of our proposed approach is to train the target-oriented features that spread out the distributions of patch features according to the variations in normal samples, and gathers the similar features.

\subsection{Overall structure}
As shown in Fig. \ref{fig:overall_structure}, our framework consists of the training and the inference phases.
In the training phase, we first collect the feature map at layer $l$, $\Phi_l\left(x\right)\in\mathbb{R}^{C\times H\times W}$ for each input $x$ in the training data using the pre-trained CNN model.

The feature maps have different spatial resolutions at the feature hierarchy of the CNN, so they are interpolated to have the same resolution before being concatenated. Patch-level features $\mathcal{P}(x,h,w)\in\mathbb{R}^{C'}$\footnote{$C$ and $C'$ can be different according to the aggregation.} then generated by aggregating the feature vectors of the neighborhood within a specific patch size $s$ in the same approach employed in PatchCore \cite{roth2022towards}. Adaptive average pooling is used for the local aggregation. 

ReConPatch utilizes two networks to train representations of the patch-level features.
One of these is a network for patch-level feature representation learning, which is trained using the relaxed contrastive loss $\mathcal{L}_{RC}$ in Eq. \ref{eq:L_RC}.
The representation network is composed of a feature representation layer $f$ and the projection layer $g$ respectively. 
When computing the $\mathcal{L}_{RC}$, pseudo-labels should be provided for every pair of features.
The other network is used to calculate pairwise and contextual similarities between patch-level feature pairs.
In addition, the similarity calculation network is gradually updated by an exponential moving average (EMA) of the representation network. To distinguish the two networks, the layers in the latter network is denoted as $\bar{f}$ and $\bar{g}$ respectively.

After training the representation, the patch-level features extracted from the pre-trained CNN are transformed into target-oriented features using the feature representation layer $f$ \cite{chen2020simple}.
The representative features are selected using the coreset subsampling approach based on the greedy approximation algorithm \cite{sinha2020gan} and stored in a memory bank. 
In the inference phase, the features of a test sample are extracted using the same process as training, and the anomaly score is calculated by comparing the features with the nominal representative in the memory bank.

\subsection{Patch-level feature representation learning} \label{sec3.2}
The objective of ReConPatch is to learn target-oriented features from patch-level features, thereby enabling more effective discrimination between normal and abnormal features.
To accomplish this goal, a patch-level features representation learning approach is applied to aggregate highly similar features while repelling those with low similarity.
However, such training requires labeled pairs to indicate the degree of proximity between patch-level features.
To address this issue, we utilize the similarity between patch-level features using the pairwise similarity and the contextual similarities as pseudo-labels. The similarity is high, then the pair is pseudo-labeled as positive and vice versa.

For two arbitrary patch-level features $p_i$ and $p_j$ obtained by $\mathcal{P}(x,h,w)$, 
let the projected representation be $\bar{z}_i=\bar{g}(\bar{f}(p_i))$ and $\bar{z}_j=\bar{g}(\bar{f}(p_j))$.
The pairwise similarity between two features, $\omega_{ij}^{Pairwise}$, is then provided by
\begin{equation} 
\omega_{ij}^{Pairwise}=e^{{-\Vert \bar{z}_i-\bar{z}_j \Vert^{2}_2}/{\sigma}}
\label{eq:pairwise_similarity}
\end{equation} 
where $\sigma$ is the bandwidth of the Gaussian kernel, which can be adjusted to tune the degree of smoothing in the similarity measure \cite{Kim2021embedding,kim2022self}.

\begin{figure}
\subfloat[]{{\includegraphics[width=0.5\columnwidth]{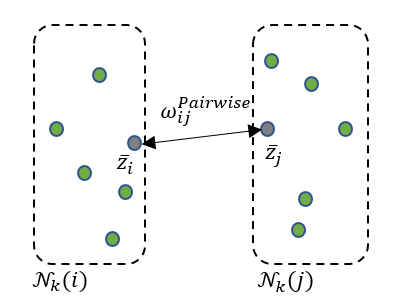}}}
\subfloat[]{{\includegraphics[width=0.5\columnwidth]{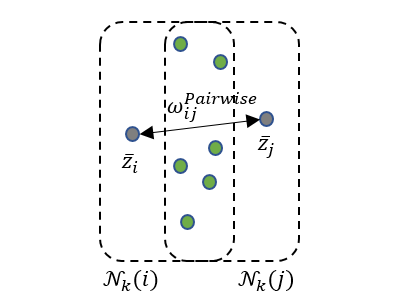}}}
\caption{Illustrative examples of similarity measures in the representation space. The pairwise similarity $\omega_{ij}^{Pairwise}$ between $\bar{z}_i$ and $\bar{z}_j$ is identical in both (a) and (b).
In (a), the $k$-nearest neighbors $\mathcal{N}_k(i)$ and $\mathcal{N}_k(j)$ do not enclose each other. Therefore, $\omega_{ij}^{Contextual}$ has a lower value, and the $\bar{z}_i$ and $\bar{z}_j$ pair should become apart.
By contrast, as $\mathcal{N}_k(i)$ and $\mathcal{N}_k(j)$ enclose each other in (b) case,
$\omega_{ij}^{Contextual}$ takes a higher value, so that $\bar{z}_i$ and $\bar{z}_j$ pair should attract each other.}
\label{fig:similarity}
\end{figure}
We note that Eq. \ref{eq:pairwise_similarity} is used to measure the Gaussian kernel similarity between $p_i$ and $p_j$, which is widely used to measure anomaly scores.
However, the pairwise similarity is insufficient to consider the relationships among groups of features.
As depicted in Fig. \ref{fig:similarity}, for example, cases (a) and (b) have the same pairwise similarity. In (a) case, $\bar{z}_i$ and $\bar{z}_j$ belong to different groups of features; therefore, they should be separated. By contrast, in (b), they belong to the same group and should be gathered.  

This leads to the simultaneous measure of contextual similarity, which consider the neighborhood of an embedding vector.
Let $k$-nearest neighborhood of the feature index $i$ is given as a set of indices,
$\mathcal{N}_k\left(i\right)=\left\{j|d_{ij}\leq d_{il}\right\}$
where $l$ is $k$-th nearest neighbor and $d_{ij}$ denotes the Euclidean distance between the two embedding vectors $(d_{ij}{=}\lVert\overline{z}_i{-}\overline{z}_j\rVert_2^2)$. 
Two patch-level features can be regarded as contextually similar if they share more nearest neighbors in common \cite{liao2022contextual}.
The contextual similarity ${\widetilde{\omega}}_{ij}^{Context}$ between two patch-level features $p_i$ and $p_j$ is then defined as
\begin{equation}
\label{eq:contextual}
\widetilde{\omega}_{ij}^{Contextual}=
\begin{cases}
{\frac{|\mathcal{N}_k(i) \cap \mathcal{N}_k(j)|}{|\mathcal{N}_k(i)|}}, \quad &\text{if }j \in \mathcal{N}_k(i)\\
0, \quad &otherwise.
\end{cases}
\end{equation}

In addition, the approach developed in this study adopts the idea of query expansion, which is widely used to improve the information retrieval, by expanding the query to the neighbors of neighbors \cite{kim2022self,liao2022contextual}. 
${\widetilde{\omega}}_{ij}^{Context}$ is redefined by averaging the similarities over the set of $k$-nearest reciprocal neighbors.
\begin{equation}
\mathcal{R}_k\left(i\right)=\left\{j\middle|j\in\mathcal{N}_k\left(i\right) \text{ and } i\in\mathcal{N}_k\left(j\right)\right\}
\end{equation}
\begin{equation}
{\hat{\omega}}_{ij}^{Contextual}=\frac{1}{\left|\mathcal{R}_{k\over2}\left(i\right)\right|}\sum_{l\in\mathcal{R}_{k\over2}\left(i\right)}{\widetilde{\omega}}_{lj}^{Contextual}.
\end{equation}
Because ${\hat{w}}_{ij}^{Context}$ is asymmetric, the contextual similarity is finally defined as the average bi-directional similarity of a pair, which is given by
\begin{equation}
\omega_{ij}^{Contextual}=\frac{1}{2}\left({\hat{\omega}}_{ij}^{Contextual}+{\hat{\omega}}_{ji}^{Contextual}\right).
\end{equation}
The final similarity between two patch-level features $p_i$ and $p_j$ is then defined as a linear combination of two similarities with a quantity $\alpha\in[0,1]$,
\begin{equation}\label{eq:weight}
\omega_{ij}=\alpha\cdot\omega_{ij}^{Pairwise}+\left(1-\alpha\right)\cdot\omega_{ij}^{Contextual}.
\end{equation}

Patch-level features do not have explicit labels because each patch image is correlated with neighboring patches. 
Moreover, the goal is to obtain unique target-oriented features rather than clearly distinguishing them.
Thus, relaxed contrastive loss \cite{Kim2021embedding} was adopted, in which inter-feature similarity is considered as pseudo-labels. Let $\delta_{ij}{=}\Vert z_i-z_j\Vert_2/({1 \over N}\Sigma_{n=1}^{N} \Vert z_i-z_n \Vert_2)$ denote the relative distance between embedding vectors in a mini-batch. The relaxed contrastive loss is given by

\begin{align}
\mathcal{L}_{RC}\!\left(z\right)\!{=}\frac{1}{N}\!\sum_{i=1}^{N}\!\sum_{j=1}^{N}{\omega_{ij}\delta_{ij}^2}
{+}{\left(1{-}\omega_{ij}\right)\!\max(m{-}\delta_{ij},\!0)^2}
\label{eq:L_RC}
\end{align}
where $z$ is the embedding vectors inferred by $g\left(f(p)\right)$, $N$ is the number of instances in a mini-batch, and $m$ is the repelling margin. $\omega_{ij}$ in Eq. \ref{eq:L_RC} determines the weights of the attracting and repelling loss terms.

While representation learning networks $f$ and $g$ are trained with relaxed contrastive loss, the similarity calculation network $\bar{f}$ and $\bar{g}$ are slowly updated with an the EMA of the parameters in $f$ and $g$ respectively. Fast training of the similarity calculation network reduces the consistency of the relationships between the patch-level features, leading to unstable training. Let $\theta_{\bar{f}, \bar{g}}$ be the parameters of the similarity calculation network and $\theta_{f, g}$ be the parameters of the representation learning network. $\theta_{\bar{f}, \bar{g}}$ is then updated by
\begin{equation} \theta_{\bar{f}, \bar{g}}\gets\gamma\cdot\theta_{\bar{f}, \bar{g}}+\left(1-\gamma\right)\cdot\theta_{f, g} \end{equation}
where $\gamma$ is the hyper-parameter that adjusts the rate of momentum update.

\subsection{Anomaly detection with ReConPatch}
Anomaly scores are calculated in the same manner as in the case of PatchCore \cite{roth2022towards}. 
After training, the coreset is subsampled from the newly trained feature representation $f(\cdot)$ using the greedy approximation algorithm \cite{sinha2020gan} and stored in memory bank $\mathcal{M}$.
The coreset takes a role of the representative feature, which is used to compute the anomaly score.
The pixel-wise anomaly score is then obtained by calculating the distance between the representation layer output, $f(p_t)$, and its nearest coreset $r^*$ within the memory bank,
\begin{gather}
r^*=\argmin_{r \in \mathcal{M}}\mathcal{D}(f(p_t),r),\\
s_t=\left(1{-}\frac{e^{s_{t}'}}{\sum_{r' \in \mathcal{N}_{b}(r^*)} e^{\mathcal{D}(f(p_t),r')}}\right)\mathcal{D}(f(p_t),r^*),
\end{gather}
where $\mathcal{N}_{b}(r^*)$ is the set of $b$-nearest neighbors of $r^*$ in the memory bank.
In addition, the image-wise anomaly score is computed as the maximum score over the anomaly scores calculated for every patch feature in the image.

The accuracy of anomaly detection can be further improved by score-level ensemble from multiple models. 
To compensate different score distribution of each model, we normalize each score using the modified z-score \cite{agga2019zscore}, normalization is necessary to evenly fuse the score levels of each model. The anomaly score is normalized to the modified z-score \cite{agga2019zscore},  defined as
\begin{equation} \label{zscore}
{\bar{s}_{t}}=\frac{s_t-\widetilde{s}}{\beta\cdot MAD},
\end{equation}
where $\widetilde{s}$ and $MAD$ are the median value of the anomaly scores and the Mean Absolute Deviation over the entire dataset for training. $\beta$ is a constant scale factor, which is set to 1.4826 in our method, assuming the anomaly score is normally distributed.

\section{Experiments and analysis}
\subsection{Experimental setup}
\textbf{Dataset}
In this study, we used the MVTec AD \cite{bergmann2019mvtec} dataset and BTAD \cite{mishra2021vt} dataset for our experiments. MVTec AD dataset is widely used as an industrial anomaly detection benchmark. It consists of 15 categories, with 3,629 training images and 1,725 test images. BTAD dataset is composed of RGB images representing three distinct industrial products. The dataset consists of 1,799 images for training and 741 images for testing. 
The training dataset includes only normal images, whereas the test dataset includes both normal and anomalous images. Each category in the test dataset has labels for normal and abnormal images, and anomaly ground truth mask labels for segmentation evaluation.

\textbf{Metrics}
To evaluate the performance of our proposed model, anomaly detection and segmentation performance is compared using the area under the receiver operation characteristic (AUROC)  curve metric, following \cite{cohen2020sub, defard2021padim, lee2022cfa, roth2022towards}.
For detection performance evaluation, we measure the image-level AUROC by using the model output anomaly score and the normal/abnormal labels of the test dataset. For segmentation, we measures the pixel-level AUROC using the anomaly scores obtained from the model output for all pixels and the anomaly ground truth mask labels.

\begin{table}[]
\centering
\begin{tabular}{l|ccc}
\specialrule{.05em}{0em}{0em} 
\hline
 \multicolumn{1}{c|}{Method}   & Ours-25\% & Ours-10\% & Ours-1\% \\
\hline
Detection    & 99.24           & 99.27                    & \textbf{99.49}          \\
Segmentation & 98.01           & \textbf{98.07}           & \textbf{98.07}          \\
\specialrule{.05em}{0em}{0em} 
\hline
\end{tabular}
\caption{Ablation study results on the coreset subsampling percentage for our proposed ReConPatch model with a WideResNet-50 backbone on the MVTec AD dataset.}
\label{table:coreset_sampling}
\end{table}

\begin{table}[]
\centering
\begin{tabular}{l|ccccc}
\specialrule{.05em}{0em}{0em} 
\hline
\multicolumn{1}{c|}{Dimension} & 1024  & 512   & 256   & 128   & 64    \\
\hline
PatchCore                     & \textbf{99.1}          & 98.66 & 98.45 & 98.54 & 97.75 \\
ReConPatch                  & 99.49 & \textbf{99.56} & 99.53 & 99.52 & 99.14 \\
\specialrule{.05em}{0em}{0em} 
\hline
\end{tabular}
\caption{Ablation study results for the $f$ layer dimension on the MVTec AD dataset using PatchCore\cite{roth2022towards} and proposed ReConPatch model with a WideResNet-50 backbone.}
\label{table:dimension_reduction}
\end{table}

\begin{table}[]
\centering
\begin{tabular}{c|c|c}
\specialrule{.05em}{0em}{0em} 
\hline
Metric       & Detection              & Segmentation             \\
\hline
\multicolumn{3}{l}{WRN-50, $s=3$, 512 dim, layer (2+3), Imagesize 224}    \\
\hline
AUROC                & 99.56                  & 98.07                    \\
\hline
\multicolumn{3}{l}{WRN-50, $s=5$, 512 dim, layer (2+3), Imagesize 224}    \\
\hline
AUROC                & 98.84                  & 97.82                    \\
\hline
\multicolumn{3}{l}{WRN-50, $s=5$, 512 dim, layer (1+2+3), Imagesize 224}  \\
\hline
AUROC                & 98.7                   & \textbf{98.18}                    \\
\specialrule{.05em}{0em}{0em} 
\hline
\end{tabular}
\caption{Ablation study results with adding more hierarchy levels and larger patch size for our proposed ReConPatch model on the MVTec AD dataset.}
\label{table:ablation_reconpatch}
\end{table}

\begin{table}[]
\centering
\resizebox{\columnwidth}{!}{%
\begin{tabular}{c|@{}c@{}|ccc}
\specialrule{.05em}{0em}{0em} 
\hline
Method & \begin{tabular}{c} \multicolumn{1}{r}{Class $\rightarrow$}\\\multicolumn{1}{l}{$\downarrow$ Aug. Method}\end{tabular} & Object   & Texture  & Average  \\
\specialrule{.05em}{0em}{0em} \hline
      & w/o Aug.      & 99.17 & 98.96 & 99.10  \\
PatchCore & w/ Aug.       & 94.86 & 96.09 & 95.48 \\
       & Diff.       & 9.94  & 2.87  & 3.62  \\\hline
       & w/o Aug.      & 99.44 & 99.81 & 99.56 \\
ReConPatch & w/ Aug.       & 97.65 & 99.47 & 98.56 \\
       & Diff.       & 1.79  & 0.34  & 1.00  \\
\specialrule{.05em}{0em}{0em} \hline
\end{tabular}
}
\caption{Ablation study results for data augmentation on MVTec AD dataset using PatchCore\cite{roth2022towards} and proposed ReConPatch.}
\label{table:aug_performance}
\end{table}

\textbf{Implementation details.}
For the single model, ImageNet pre-trained WideResNet-50 \cite{zagoruyko2016wide} is employed as the feature extractor. The $f$ layer output size is set to 512, and the coreset subsampling percentage is set to 1\%. Our proposed ReConPatch is trained for 120 epochs per each category. Without specific instructions, hierarchy levels\footnote{Hierarchy levels denote residual blocks in WideResNet architecture, which is same in \cite{roth2022towards}.} 2 and 3 are used with a patch size of $s=3$ to generate the patch-level features.
Particularly for the segmentation evaluation in Table \ref{table:performance}, hierarchy levels 1, 2, and 3 were used with a patch size of $s=5$, which is identified as the best performance through the ablation study in section \ref{ablation}. In addition, for the comparison with PNI \cite{bae2023pni} using WideResNet-101, hierarchy levels 2 and 3 were used with a patch size of $s=5$.

For the ensemble model, ImageNet pre-trained WideResNet-101 \cite{zagoruyko2016wide}, ResNext-101 \cite{xie2017aggregated}, and DenseNet-201 \cite{huang2017densely} are used as feature extractors for comparison with the PatchCore \cite{roth2022towards}. The $f$ layer output size was set to 384, and we applied a coreset subsampling with percentage of 1\% to all models in the ensemble. We trained ReConPatch for 60 epochs for each category. Hierarchy levels 2 and 3 were used for feature extraction in each model, and a patch size of $s=3$ was applied to generate the patch-level features. Furthermore, to compare with PNI \cite{bae2023pni} using 480$\times$480 image size, different parameters were applied. The $f$ layer output size was set to 512, and a patch size of $s=5$ was used. In this case, we trained each category for 120 epochs. ReConPatch was trained using AdamP \cite{heo2020adamp} optimizer with a cosine annealing \cite{loshchilov2016sgdr} scheduler. The learning rate was set to 1e-5 for a single model and 1e-6 for the ensemble model, with a weight decay of 1e-2. In the models using a 480$\times$480 image size, the learning rate was specifically set to 1e-6. We provide the hyperparameter setup in Appendix B.

\begin{table*}[]
\centering
\resizebox{\textwidth}{!}{%
\begin{tabular}{l|cc|cccccc}
\specialrule{.05em}{0em}{0em} 
\hline
Backbone                    & \multicolumn{2}{c|}{WRN-101}  & \multicolumn{6}{c}{WRN-50}        \\
\hline
Image size                  & 480$\times$480 & 480$\times$480 & 256$\times$256 & 224$\times$224 & 224$\times$224 & 224$\times$224 & 224$\times$224 & 224$\times$224 \\
\hline
$\downarrow$ Class\textbackslash{}Method $\rightarrow$ & \begin{tabular}[c]{@{}c@{}}PNI \cite{bae2023pni}\\ (w/ refine)\end{tabular} & \textbf{Ours} & CFLOW-AD \cite{gudovskiy2022cflow} & SPADE \cite{cohen2020sub} & PaDiM \cite{defard2021padim} &  PatchCore \cite{roth2022towards}    & CFA \cite{lee2022cfa} & \textbf{Ours}  \\
\hline
Bottle       & (\textbf{100}, \textbf{98.87})     & (\textbf{100}, 98.78)     & (\textbf{100}, \textbf{98.76})   & (-, 98.4) & (-, 98.3)    & (\textbf{100}, 98.6)     & (\textbf{100}, -)       & (\textbf{100}, 98.2)    \\
Cable        & (\textbf{99.76}, \textbf{99.1})    & (99.66, 98.86)            & (97.59, 97.64)                   & (-, 97.2) & (-, 96.7)    & (99.5, 98.4)             & (99.8, -)      & (\textbf{99.83}, \textbf{99.3})  \\
Capsule      & (99.72, \textbf{99.34})            & (\textbf{99.76}, 99.24)   & (97.68, 98.98)            & (-, \textbf{99}) & (-, 98.5)    & (98.1, 98.8)             & (97.3, -)               & (\textbf{98.8}, 97.61)  \\
Hazelnut     & (\textbf{100}, \textbf{99.37})     & (\textbf{100}, 99.07)     & (99.98, 98.82)          & (-, \textbf{99.1}) & (-, 98.2)    & (\textbf{100}, 98.7)     & (\textbf{100}, -)       & (\textbf{100}, 98.94)   \\
Metal nut    & (\textbf{100}, \textbf{99.29})     & (\textbf{100}, \textbf{99.29})   & (99.26, \textbf{98.56})   & (-, 98.1) & (-, 97.2)    & (\textbf{100}, 98.4)     & (\textbf{100}, -)       & (\textbf{100}, 95.76)   \\
Pill         & (\textbf{96.89}, \textbf{99.03})   & (96.21, 98.66)            & (96.82, \textbf{98.95})          & (-, 96.5) & (-, 95.7)    & (96.6, 97.4)             & (\textbf{97.9}, -)      & (97.49, 95.35)          \\
Screw        & (99.51, \textbf{99.6})             & (\textbf{99.84}, 99.59)   & (91.89, 98.1)        & (-, 98.9) & (-, 98.5)                & (98.1, \textbf{99.4})    & (97.3, -)               & (\textbf{98.52}, 98.79) \\
Toothbrush   & (99.72, 99.09)       & (\textbf{100}, \textbf{99.16})          & (99.65, 98.56)       & (-, 97.9) & (-, 98.8)    & (\textbf{100}, 98.7)     & (\textbf{100}, -)       & (\textbf{100}, \textbf{98.88})      \\
Transistor   & (\textbf{100}, \textbf{98.04})     & (\textbf{100}, 96.18)     & (95.21, 93.28)       & (-, 94.1) & (-, 97.5)    & (\textbf{100}, 96.3)     & (\textbf{100}, -)       & (\textbf{100}, \textbf{99.65})      \\
Zipper       & (99.87, \textbf{99.43})           & (\textbf{99.89}, 99.25)    & (98.48, 98.41)                   & (-, 96.5) & (-, 98.5)    & (99.4, \textbf{98.8})    & (99.6, -)               & (\textbf{99.76}, 98.56) \\
\hline
\multicolumn{1}{c|}{Object classes} & (\textbf{99.55}, \textbf{99.12})        & (99.54, 98.81)        & (97.66, 98.01)  & (-, 97.57)  & (-, 97.79)   & (99.17, \textbf{98.35})    & (99.19, -)      & (\textbf{99.44}, 98.1) \\
\hline 
Carpet       & (\textbf{100}, \textbf{99.4})     & (\textbf{100}, 99.29)      & (98.73, \textbf{99.23})          & (-, 97.5) & (-, 99.1)    & (98.7, 99)               & (97.3, -)               & (\textbf{99.6}, 98.75)  \\
Grid         & (98.41, \textbf{99.2})            & (\textbf{99.5}, 98.73)     & (99.6, 96.89)        & (-, 93.7) & (-, 97.3)    & (98.2, 98.7)             & (99.2, -)               & (\textbf{100}, \textbf{99.04})      \\
Leather      & (\textbf{100}, \textbf{99.56})    & (\textbf{100}, 99.48)      & (\textbf{100}, \textbf{99.61})   & (-, 97.6) & (-, 99.2)    & (\textbf{100}, 99.3)     & (\textbf{100}, -)       & (\textbf{100}, 96.02)   \\
Tile         & (\textbf{100}, \textbf{98.4})     & (\textbf{100}, 97.15)      & (\textbf{99.88}, 97.71)          & (-, 87.4) & (-, 94.1)    & (98.7, 95.6)             & (99.4, -)               & (99.78, \textbf{98.92}) \\
Wood         & (\textbf{99.56}, \textbf{97.04})  & (99.47, 95.16)             & (99.12, 94.49)                   & (-, 88.5) & (-, 94.9)    & (99.2, 95)               & (\textbf{99.7}, -)      & (99.65, \textbf{98.9})  \\
\hline
\multicolumn{1}{c|}{Texture classes} & (99.59, \textbf{98.72})   & (\textbf{99.79}, 97.96)     & (99.47, 97.59)   & (-, 92.94)& (-, 96.92)   & (98.96, 97.52)           & (99.12, -)      & (\textbf{99.81}, \textbf{98.33})  \\
\hline
\multicolumn{1}{c|}{Average} & (99.56, \textbf{98.98})   & (\textbf{99.62}, 98.53)             & (98.26, 97.87)  & (85.5, 96)& (95.3, 97.5) & (99.1, 98.1)         & (99.2, \textbf{98.2})      & (\textbf{99.56}, 98.18)  \\
\specialrule{.05em}{0em}{0em} 
\hline
\end{tabular}%
}
\caption{Anomaly detection and segmentation performance on the MVTec AD dataset. (image-level AUROC, pixel-level AUROC)}
\label{table:performance}
\end{table*}

\subsection{Ablation study} \label{ablation}
In this study, we aim to investigate the optimal configuration of ReConPatch through ablation studies. The first ablation was performed to determine the optimal coreset subsampling percentage. To this end, we compared anomaly detection and segmentation AUROC metrics using three subsampling percentages: 25\%, 10\%, and 1\%, which were the same percentages used in PatchCore \cite{roth2022towards}. The pre-trained WideResNet-50 \cite{zagoruyko2016wide} backbone was used as the baseline for this experiment and the output dimension of the $f$ layer is set to 1024. The results are presented in Table \ref{table:coreset_sampling}. We observe that the subsampling percentage of 1\% provides the best performance.
In addition, experiments to analyze the performance according to the feature dimension were performed by changing various output dimension of the $f$ layer (1024, 512, 256, 128, and 64). The experiments were conducted with coreset subsampling set to 1\%. The results are presented in Table \ref{table:dimension_reduction}, indicating that the highest performance was achieved with the dimension of 512. We note that even with 64 dimension, ReConPatch outperforms PatchCore with 1024, which supports the dimension reduction capability of our method.

Table \ref{table:ablation_reconpatch} shows the results of an ablation study using more hierarchy levels and larger patch size on the MVTec AD \cite{bergmann2019mvtec} dataset with our proposed ReConPatch model. This study aims to improve segmentation performance by utilizing more diverse and coarse information on the patch features. The results indicates that when the patch size is increased to 5 and hierarchy levels 1, 2, and 3 are used,  the segmentation performance increased up to 98.18\% with small decrease in detection performance.

Real-world scenarios can present a variety of environmental conditions that can affect the quality of images.
These conditions may include geometric changes, lighting changes, defocusing, and other factors that can impact the accuracy and reliability of image data. Table \ref{table:aug_performance} shows that ReConPatch is robust to these environmental changes by learning patch-level feature representations. To simulate real-world scenarios, we randomly applied rotation, translation, color jitter (brightness and contrast), and Gaussian blur. While PatchCore’s image-level AUROC decreased to 3.62 under these conditions, ReConPatch’s only slightly decreased to 1.0.

\begin{table}[t]
\centering
\resizebox{\columnwidth}{!}{%
\begin{tabular}{l|cc|cc}
\specialrule{.05em}{0em}{0em} 
\hline
\begin{tabular}[c]{@{}l@{}}Ensemble\\ Backbone\end{tabular} & \multicolumn{4}{c}{WRN-101 \& RNext-101 \& DenseN-201}                          \\
\hline
Image size           & 480$\times$480      & 480$\times$480            & 320$\times$320             & 320$\times$320 \\
\hline
Method               & \begin{tabular}[c]{@{}c@{}}PNI \cite{bae2023pni} \\ (w/ refine)\end{tabular} & \textbf{Ours} & \begin{tabular}[c]{@{}c@{}}PatchCore \\\cite{roth2022towards}\end{tabular} & \textbf{Ours}    \\
\hline
Detection                     & 99.63               & \textbf{99.72}                & 99.6      & \textbf{99.67}   \\
Segmentation                  & \textbf{99.06}      & 98.67                & 98.2      & \textbf{98.36}   \\
\specialrule{.05em}{0em}{0em} 
\hline
\end{tabular}%
}
\caption{Comparison of ensemble model anomaly detection (image-level AUROC) and segmentation (pixel-level AUROC) performance on the MVTec AD dataset.}
\label{table:ensemble_result}
\end{table}

\begin{table*}[]
\resizebox{\textwidth}{!}{%
\centering
\begin{tabular}{c|ccccccccccc}
\specialrule{.05em}{0em}{0em} 
\hline
Class & VT-ADL \cite{mishra2021vt} & SPADE \cite{cohen2020sub} & PaDiM \cite{defard2021padim} & FastFlow \cite{yu2021fastflow} & PyramidFlow \cite{lei2023pyramidflow} &  CFA \cite{lee2022cfa}  & RD4AD \cite{deng2022anomaly} & RD++ \cite{tien2023revisiting} & PNI \cite{bae2023pni} & PatchCore \cite{roth2022towards} & \textbf{Ours}  \\
\hline
1      & (97.6, \textbf{99})     & (91.4, 97.3)         & (99.8, 97)      & (99.4, 97.1)     & (\textbf{100}, 97.4)     & (98.1, 95.9)   & (96.3, 96.6)      & (96.8, 96.2)   & (-, 97.4) & (98, 96.9) & (99.7, 96.8) \\
2      & (71, 94)                & (71.4, 94.4)         & (82, 96)        & (82.4, 93.6)     & (88.2, \textbf{97.6})    & (85.5, 96)     & (86.6, 96.7)      & (\textbf{90.1}, 96.4)   & (-, 97)  & (81.6, 95.8) & (87.7, 96.6) \\
3      & (82.6, 77)              & (99.9, 99.1)         & (99.4, 98.8)    & (91.1, 98.3)     & (99.3, 98.1)             & (99, 98.6)     & (\textbf{100}, \textbf{99.7})   & (\textbf{100}, \textbf{99.7})  & (-, 99) & (99.8, 99.1) & (\textbf{100}, 99) \\
\hline
Avg. & (83.7, 90)      & (87.6, 96.9)     & (93.7, 97.3)  & (91, 96.3) & (\textbf{95.8}, 97.7) &  (94.2, 96.8)   & (94.3, 97.7)   & (95.6, 97.4)      & (-, \textbf{97.8})  & (93.1, 97.3) & (\textbf{95.8}, 97.5)  \\
\specialrule{.05em}{0em}{0em} 
\hline
\end{tabular}%
}
\caption{Anomaly detection and segmentation performance on the BTAD \cite{mishra2021vt} dataset. (image-level AUROC, pixel-level AUROC)}
\label{table:performance_btad}
\end{table*}

\subsection{Anomaly detection on MVTec AD}

In this section, we evaluate the anomaly detection performance of our proposed method on the MVTec AD dataset by comparing it with previous works that used the same pre-trained model and image size \cite{cohen2020sub, defard2021padim, lee2022cfa, roth2022towards}. 
We also include the performance of concurrent methods PNI \cite{bae2023pni} and CFLOW-AD \cite{gudovskiy2022cflow} in Tables \ref{table:performance}.
In case of PNI \cite{bae2023pni}, a WideResNet-101 model with an image size of 480$\times$480 was used. To improve its performance, a refinement network was included, which was trained in a supervised manner using artificially created defect dataset. For CFLOW-AD \cite{gudovskiy2022cflow}, a WideResNet-50 model with an image size of 256$\times$256 is used. The evaluation results used in CFLOW-AD were the best performances obtained for each category when using the image size of 256$\times$256. 

For the single-model performance comparison, we performed the same pre-processing as described in previous work \cite{cohen2020sub, defard2021padim, lee2022cfa, roth2022towards}. Specifically, we resized each image to 256$\times$256 and then center-cropped to 224$\times$224. For the ensemble model, the same pre-processing was used as in \cite{roth2022towards}, each image was resized to 366$\times$366 and then center-cropped to 320$\times$320. In addition, to compare with PNI \cite{bae2023pni}, we resized each image to 512$\times$512 and then center-cropped to 480$\times$480. No data augmentation was applied to any category. 

The performance of the ReConPatch in Tables \ref{table:performance} was obtained using 1\% coreset subsampling and $f$ layer dimensions of 512, which is determined according to Table \ref{table:dimension_reduction}.
Table \ref{table:performance} compares the anomaly detection and segmentation performance of a single model for each category of the MVTec AD \cite{bergmann2019mvtec} dataset, evaluated with image-level AUROC. Our proposed ReConPatch achieved an image-level AUROC of 99.56\%, which outperformed CFA \cite{lee2022cfa} (at 99.3\%). Furthermore, ReConPatch provided higher performance than the state-of-the-art PNI \cite{bae2023pni} with WideResNet-101 \cite{zagoruyko2016wide}, which achieved the performance of 99.62\%.

Our proposed approach focused on improving the anomaly detection performance. As a result, the segmentation performance may not be as high as its detection performance. However, we achieved a higher performance of 98.18\% compared to PatchCore \cite{roth2022towards}, indicating that the addition of ReConPatch feature in the $f$ layer contributed to the improved segmentation performance.
\begin{figure}[t]
\centering
\includegraphics[width=\linewidth]{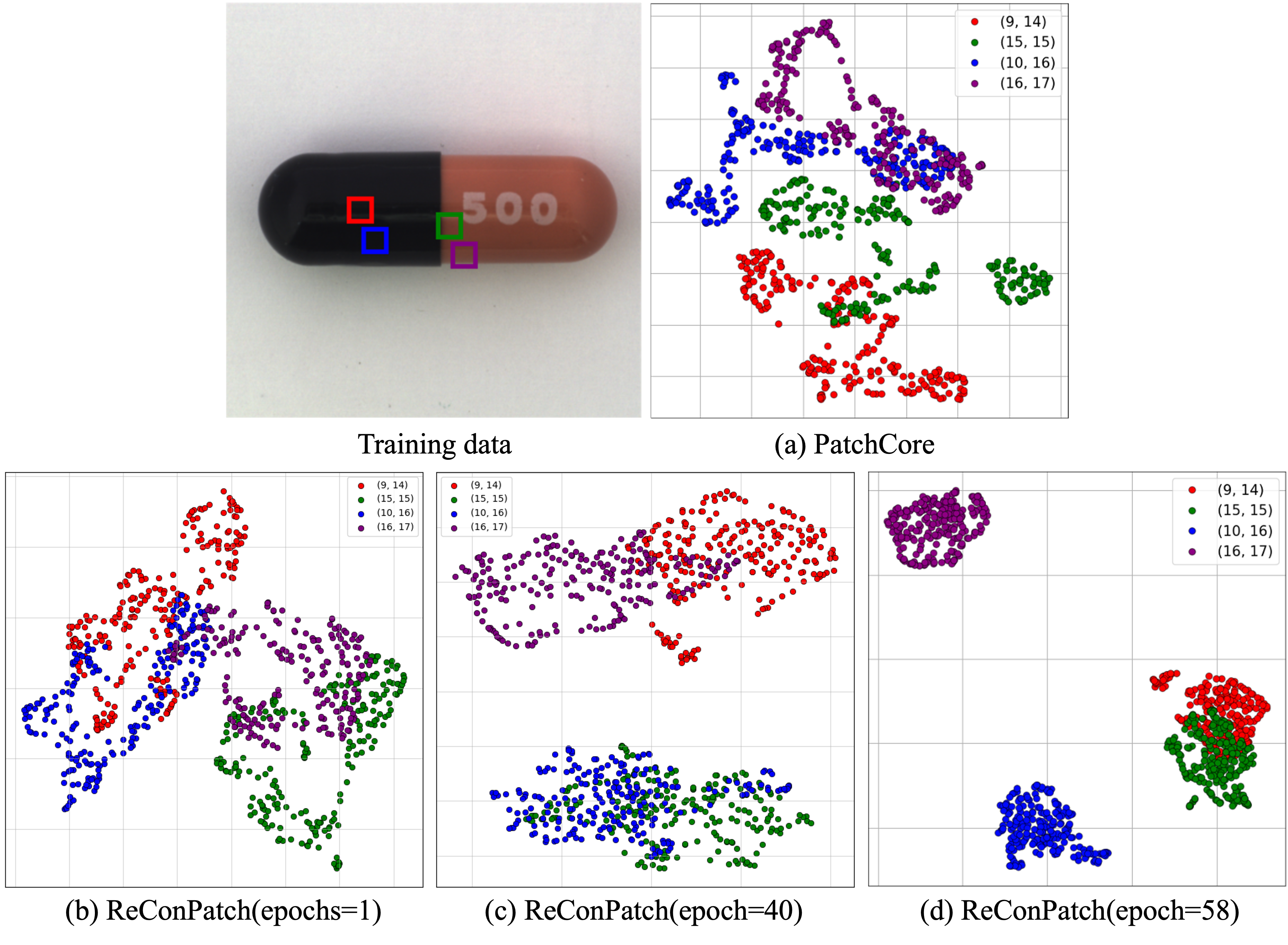}
\caption{An illustrative comparison of features mapped by (a) PatchCore and (b) (c) (d) ReConPatch using the MVTec AD dataset. The scatter plot describes the feature space of each method, colored according to the pixel position. 
}
\label{fig:feature_map}
\end{figure}

\begin{figure}[t]
\centering
\includegraphics[width=0.9\linewidth]{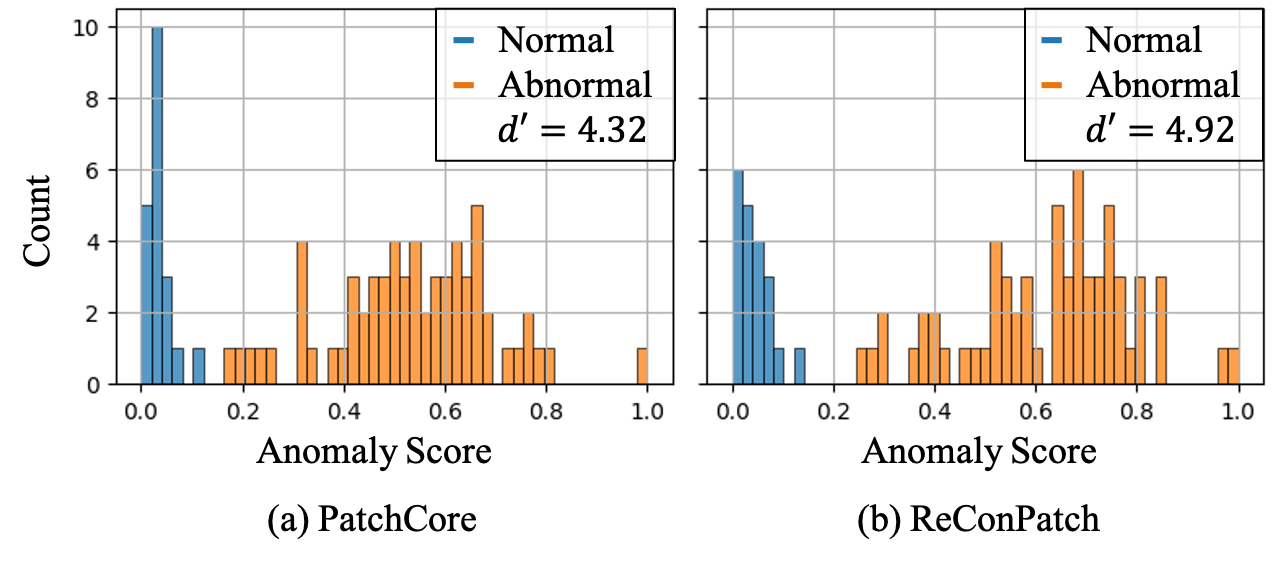}
\caption{The histogram of the anomaly score of the normal and abnormal data for the bottle class. ReConPatch shows high discriminability, as shown in $d'$ measure.
}
\label{fig:histogram}
\end{figure}

\begin{figure*}
    \centering
    \includegraphics[width=0.9\textwidth]{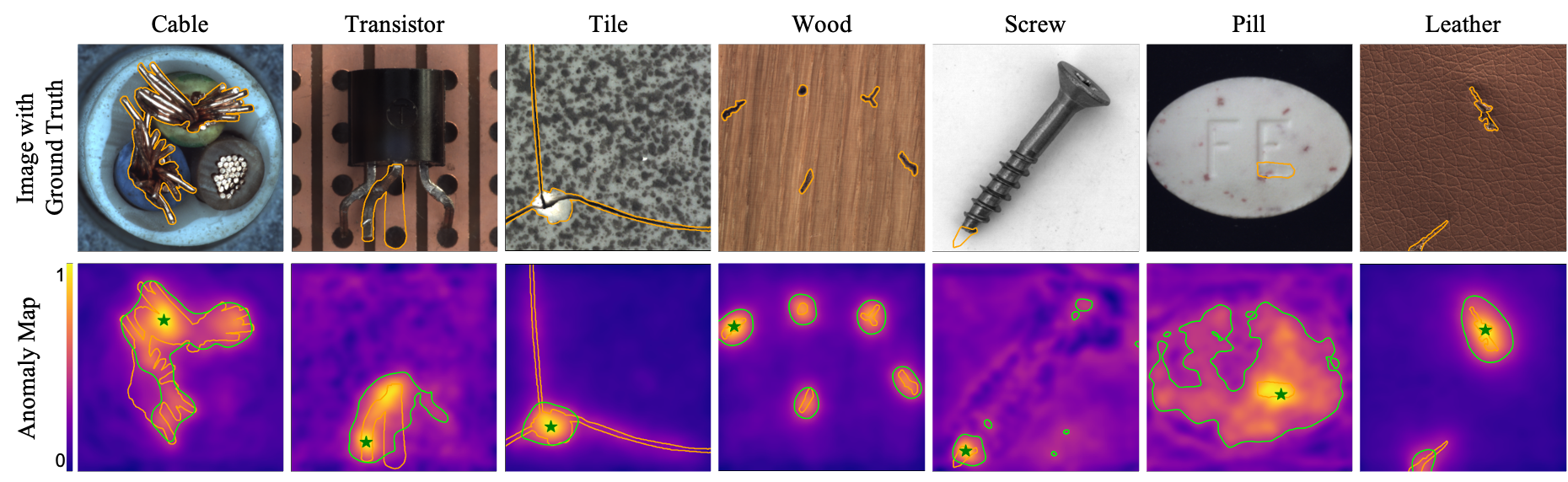}
    \caption{Examples of images with anomalies (top) and measured anomaly score maps (bottom) on MVTec AD dataset. The {\color{orange}orange line} depicts the ground truth of the anomalies and the {\color{ForestGreen}green line} depicts thresholds optimizing F1 scores of anomaly segmentation. The {\color{ForestGreen}green star} indicates the maximal location of the anomaly score in the heatmap.}
    \label{fig:example_segmentation}
\end{figure*}

Table \ref{table:ensemble_result} presents the performance of our ensemble model, which was evaluated using the modified z-score in Eq. \ref{zscore} for each output from WideResNet-101 \cite{zagoruyko2016wide}, ResNext-101 \cite{xie2017aggregated}, and DenseNet-201 \cite{huang2017densely} models. Our model achieved state-of-the-art performance in anomaly detection task with AUROC of 99.72\% on the MVTec AD dataset using an image size of 480$\times$480. 
We note that our model still outperforms the PNI \cite{bae2023pni} using a smaller image size of 320$\times$320, achieving an AUROC of 99.67\% compared to AUROC of 99.63\%
Furthermore, we outperformed PatchCore \cite{roth2022towards} in terms of anomaly segmentation performance, with an improved performance of 98.36\% AUROC.

\subsection{Anomaly detection on BTAD}

To verify the capability of anomaly detection and segmentation in other dataset, we compare the performance of our model with contemporary methods using BTAD dataset \cite{mishra2021vt}. For BTAD dataset, we use the pre-trained WideResNet-101 model as a feature extractor and image size of 480$\times$480 for ReConPatch, which achieve our best performance.
Table \ref{table:performance_btad} shows the image-level AUROC and the pixel-level AUROC on BTAD dataset.
Our model achieves a state-of-the-art performance in anomaly detection, with an AUROC of 95.8\%. Furthermore, in anomaly segmentation, our model outperforms PatchCore \cite{roth2022towards} with a higher AUROC of 97.5\%.


\subsection{Qualitative analysis}

To assess the impact of ReConPatch learning on the feature space, we contrast the feature space of PatchCore and ReConPatch using the MVTec AD dataset. Our visualization, depicted in Figure \ref{fig:feature_map}, employs UMAP \cite{mcinnes2018umap} for effective 2D representation of high-dimensional patch features, with color coding indicating spatial positions. The visualization attests that ReConPatch's training encourages proximity of features with similar positions. Building on findings in prior research \cite{bae2023pni,gudovskiy2022cflow}, which demonstrated the value of positional information, we hypothesize that ReConPatch's performance enhancement arises from implicit positional information learning. We also visualize the feature map along the training, which indicates the features are trained to map similar position to be gather.

ReConPatch's reconfigured feature space yields more distinct histogram distributions of image-level anomaly scores compared to PatchCore. In Figure \ref{fig:histogram}, we observe this effect on the MVTec AD dataset's bottle class. ReConPatch compresses the score distribution for normal data while pushing the abnormal data's distribution further from the normal one, a contrast to PatchCore \cite{roth2022towards}. We gauge the distribution separability using the $d'$ discriminability index \cite{simpson1973best} between normal and abnormal data:
\begin{align}
d' = \frac{|\mu_{abnormal} - \mu_{normal}|}{\sqrt{(\sigma_{abnormal}^2 + \sigma_{normal}^2)/2}}.
\end{align}
Here, the patch features mirror those of locally aware patch features in PatchCore. ReConPatch, as detailed in Section \ref{sec3.2}, leverages target-oriented features through patch-level representation training, enhancing discrimination between normal and abnormal attributes. Performance-wise (Table \ref{table:performance}), ReConPatch achieves an image-level AUROC of 99.56

We present anomaly score maps overlaid on input images (Figure \ref{fig:example_segmentation}) with ground truth annotations. Higher values in the anomaly map indicate probable anomalies. A threshold optimized via F1 scores governs the green line. Our analysis focuses on 4 superior classes (cable, transistor, tile, wood) and 3 inferior classes (metal nut, pill, leather). Despite intricate ground truth cases, ReConPatch consistently identifies anomaly locations. While inferior class anomaly maps may exhibit noise, the green star pinpointing maximal anomaly score aligns with ground truth anomalies, affirming our method's robust performance in anomaly detection.


\section{Conclusion}
In this paper, we introduce the ReConPatch to learn a target-oriented representation space, which can effectively distinguish the anomalies from the normal dataset. ReConPatch effectively trains the representation by applying the metric learning with softly guided by the similarity over the nominal features.
Applying the contrastive learning with two similarity based pseudo soft labels, ReConPatch shows the state-of-the-art performance on the MVTec anomaly detection dataset. We also provide the anomaly detection performance on the additional BTAD dataset, where ReConPatch also achieves the best performance.
We believe that ReConPatch would contribute to the improvements in anomaly detection since it shows high performance without extensive data augmentation and enables dimension reduction without significant loss of performance.
Furthermore, we expect to improve the performance in the pixel-level abnormal detection by considering the correlation among the neighboring features.

\clearpage
{\small
\bibliographystyle{ieee_fullname}
\bibliography{egbib}
}

\clearpage
\onecolumn

\begin{centering}
\section*{\Large{Supplementary:\\ReConPatch : Contrastive Patch Representation Learning for Industrial Anomaly Detection}}
\end{centering}
\vspace{12pt}

\setcounter{equation}{0}
\setcounter{figure}{0}
\setcounter{table}{0}
\setcounter{page}{1}
\makeatletter
\renewcommand{\theequation}{S\arabic{equation}}
\renewcommand{\thefigure}{S\arabic{figure}}
\renewcommand{\thetable}{S\arabic{table}}

\appendix
\section{Implementation details}\label{appendix_section:implementation_details}
ReConPatch was implemented with Python 3.7 \cite{python} and PyTorch 1.9 \cite{pytorch}.
Experiments are run on Nvidia GeForce GTX 3090 GPU.
We used ImageNet-pretrained models from \texttt{torchvision} \cite{torchvision2016} and the PyTorch Image Models repository \cite{rw2019timm}. By default, following \cite{cohen2020sub} and \cite{defard2021padim}, ReConPatch uses WideResNet50-backbone and WideResNet101-backbone\cite{zagoruyko2016wide} for direct comparability. Patch-level features are taken from feature map aggregation of the final outputs in $f$ layer of ReConPatch. We use \texttt{faiss} \cite{faiss} to compute all nearest neighbor retrieval and distance computations same as PatchCore\cite{roth2022towards}.

\section{Additional experiments on MVTec AD}\label{appendix_section:additional_experiments_mvtec}
This section contains the details of setting up hyper-parameters and experiments with the projection layer of ReConPatch on MVTec AD\cite{bergmann2019mvtec} dataset.
We experimented to find the optimal hyper-parameters with the following values: $k$ value of k-nearest neighborhood for calculating contextual similarity, the repelling margin $m$ in RC loss, and applying ratio $\alpha$ between pairwise and contextual similarity. 
Additionally, to verify the effectiveness of projection layer $g$, we tested our proposed method without projection layer $g$.

All experiments have the same conditions. Each images are resized to 256$\times$256 and center-cropped to 224$\times$224 and ImageNet-pretrained WideResNet50-backbone\cite{zagoruyko2016wide} from \texttt{torchvision}\cite{torchvision2016} was employed as the feature extractor.
The $f$ layer output size was set to 512, and the coreset subsampling percentage was set to 1\%. The ReConPatch was trained for 120 epochs per each class on MVTec AD \cite{bergmann2019mvtec} with AdamP\cite{heo2020adamp} optimizer with cosine annealing \cite{loshchilov2016sgdr} scheduler.

\subsection{The $k$ value of k-nearest neighbor for contextual similarity}\label{appendix_subsec:finding_optimal_k}
We tested several $k$ - nearest neighborhood value $k$ for finding the best performance of anomaly detection on MVTec AD\cite{bergmann2019mvtec} dataset. In Eq. \ref{eq:contextual}, smaller $k$ values will use features that are closer together to determine contextual similarity, while larger $k$ values will use features that are further apart in the embedding space to determine contextual similarity. To find an optimal $k$ value, we fixed other hyper-parameters $m=1$, $\alpha=0.5$. Table \ref{table:supple_k_nearest} shows the results of experiments.

\begin{table}[hbt!]
\centering
\begin{tabular}{l|cccccc}
\hline
\multicolumn{7}{l}{ReConPatch (WRS-50, 224$\times$224, $\alpha=0.5$, $m=1.0$)}  \\
\hline
\multicolumn{1}{c|}{$k$ value}
    & 3            & 5                    & 7            & 10           & 15          & 20       \\
\hline
Detection                     
    & 99.5         &\textbf{99.56}        & 99.55        & 99.54        & 99.5        & 99.49    \\
Segmentation                  
    &\textbf{98.09}& 98.07                & 98.07        & 98.07        & 98.04       & 98.07    \\
\hline
\end{tabular}
\caption{Results of anomaly detection(image-level AUROC) and segmentation(pixel-level AUROC) with various $k$-nearest neighborhood on MVTec AD dataset. ($\alpha=0.5$, $m=1.0$).}
\label{table:supple_k_nearest}
\end{table}

\subsection{The repelling margin $m$ of relaxed contrastive loss}\label{appendix_subsec:finding_optimal_m}
We tested margin value $m$ in Eq. \ref{eq:L_RC} repelling term of relaxed contrastive loss with various $m$ values to obtain the performance of anomaly detection on MVTec AD\cite{bergmann2019mvtec}. All other hyper-parameters are fixed($k=5$, $\alpha=0.5$). Table \ref{table:supple_margin} is the results of experiments. Changing the margin $m$ has no effect on performance of anomaly detection(image-level AUROC) but it does slightly affect on segmentation(pixel-level AUROC). So we use the repelling margin $m$ to 1.0.

\begin{table}[hbt!]
\centering
\begin{tabular}{l|cccc}
\hline
\multicolumn{5}{l}{ReConPatch (WRS-50, 224$\times$224, $k=5$, $\alpha=0.5$)}    \\
\hline
\multicolumn{1}{c|}{m value}
    & 0.5          & 1                    & 1.5          & 2    \\
\hline
Detection                     
    & 99.55        &\textbf{99.56}        & 99.55        & 99.5     \\
Segmentation                  
    & 98.04        &\textbf{98.07}        & 98.05        & 98.04    \\
\hline
\end{tabular}
\caption{Results of anomaly detection(image-level AUROC) and segmentation(pixel-level AUROC) with various repelling margin $m$ values of RC loss. ($k=5$, $\alpha=0.5$).}
\label{table:supple_margin}
\end{table}

\subsection{The applying ratio $\alpha$ between pairwise and contextual similarity}\label{appendix_subsec:finding_optimal_alpha}
We tested $\alpha$ in Eq. \ref{eq:weight} which the ratio of linear combination between pairwise and contextual similarity within range $[0,1]$. If $\alpha$ is 0 then only use contextual similarity, and if $\alpha$ set to 1.0 then use pairwise similarity only. Table \ref{table:supple_alpha} shows the best detection(image-level AUROC) performance when using pairwise and contextual similarity with same ratio.

\begin{table}[hbt!]
\centering
\begin{tabular}{l|ccccc}
\hline
\multicolumn{6}{l}{ReConPatch (WRS-50, 224$\times$224, $k=5$, $m=1.0$)} \\
\hline
\multicolumn{1}{c|}{$\alpha$}
    & 0.0          & 0.25         & 0.5                  & 0.75         & 1.0           \\
\hline
Detection                     
    & 99.51        & 99.51        &\textbf{99.56}        & 99.52        & 99.5          \\
Segmentation                  
    & 98.02        & 98.07        & 98.07                & 98.1         &\textbf{98.12} \\
\hline
\end{tabular}
\caption{Results of anomaly detection(image-level AUROC) and segmentation(pixel-level AUROC) with various $\alpha$ values within range $[0,1]$ of ReConPatch. ($k=5$, $m=1.0$).}
\label{table:supple_alpha}
\end{table}

\subsection{The projection layer $g$}
\begin{figure}[h]
\includegraphics[width=1\linewidth]{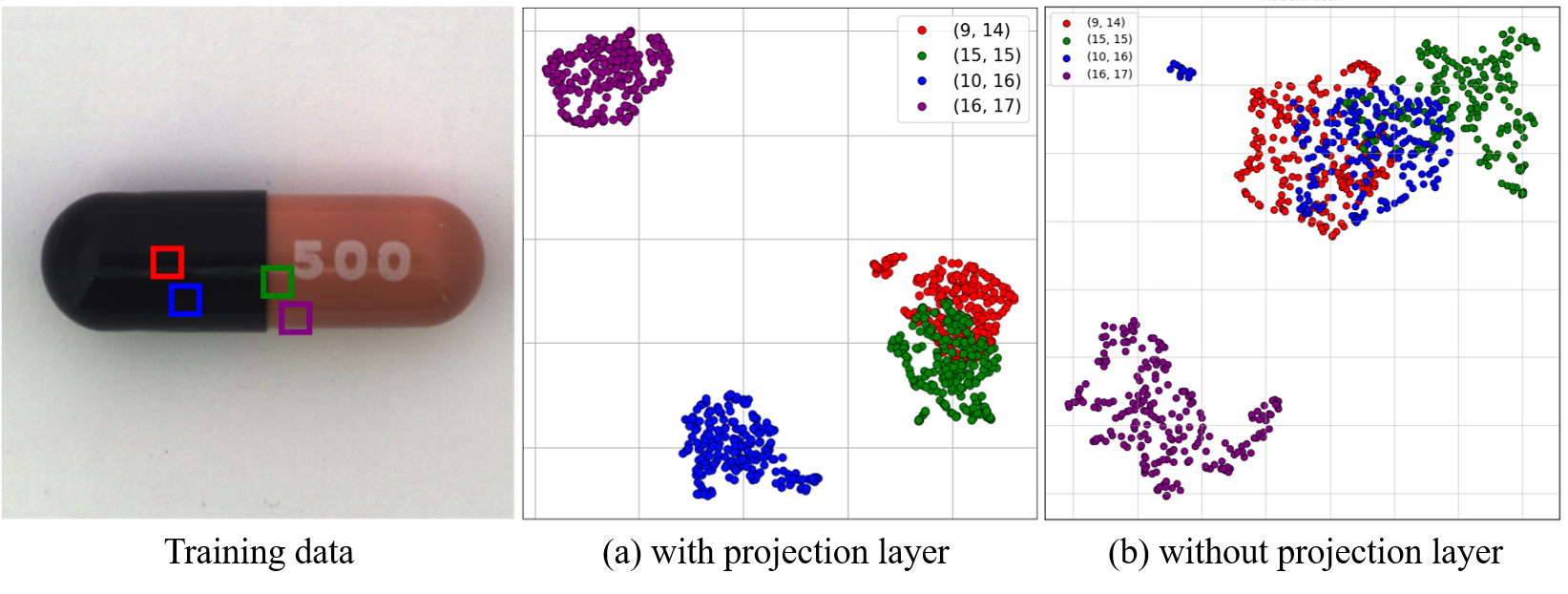}
\caption{An illustrative comparison of features mapped by ReConPatch with projection layer (a) and without projection layer (b) using the MVTec AD dataset\cite{bergmann2019mvtec}. The scatter plot describes the feature space of each method, colored according to the pixel position.}
\label{fig:supple_projection_layer}
\end{figure}

We tested with linear projection layer $g$, without projection layer $g$. Fig. \ref{fig:supple_projection_layer} describes the visualization of patch features obtained by ReConPatch which trained with projection layer $g$ and trained without projection layer. For effective visualization of high dimensional feature vectors, we map the patch features into 2-dimensional space using UMAP\cite{mcinnes2018umap}. With visualization result, we verify that the projection layer $g$ of ReConPatch encourages the features with similar position together. We can also verify that using the projection layer has better anomaly detection performance than not using the projection layer by the results in Table \ref{table:supple_projection_layer}.

\begin{table}[hbt!]
\centering
\begin{tabular}{l|cc}
\hline
\multicolumn{3}{l}{ReConPatch (WRS-50, 224$\times$224, $k=5$, $m=1.0$, $\alpha=1.0$)}\\
\hline
\multicolumn{1}{c|}{Projection layer}
    & w/o Proj. Layer    & w/ Proj. Layer    \\
\hline
Detection                     
    & 99.42              & 99.56             \\
\hline
\end{tabular}
\caption{Results of Detection AUROC with and without projection layer $g$ on the MVTec AD dataset\cite{bergmann2019mvtec} using our proposed ReConPatch model with a WideResNet-50 backbone\cite{zagoruyko2016wide}, 224$\times$224 input size, 512 dimensional $f$ layer, $k=5$, $m=1.0$, $\alpha=1.0$ and 1\% coreset sampling.}
\label{table:supple_projection_layer}
\end{table}

\section{Performance of ReConPatch}\label{appendix_section:performance}
\subsection{Applying data augmenatation}
In the industrial domain, product images are collected in a well-controlled environment, but uncontrollable change in the environmental conditions at the time of acquisition can cause variation in the image data.
This can lead to variation between normal image data, which can adversely affect the performance of industrial anomaly detection.
To verify that ReConPatch is adaptable to environmental changes, we applied various data augmentation method to evaluate its anomaly detection performance.
We randomly applied rotation($\pm5degree$), translation($\pm1\%$), color jitter(brightness and contrast $\pm30\%$), and gaussian blur($\sigma=[0.1, 1.0]$) which could occur in the real world. Table \ref{table:supple_aug_performance} shows the detail performance comparison of PatchCore\cite{roth2022towards} and ReConPatch with data augmentation on MVTec AD dataset\cite{bergmann2019mvtec}.

\begin{table*}[]
\centering
\begin{tabular}{l|ccc|ccc}
\specialrule{.05em}{0em}{0em} 
\hline
\multicolumn{1}{c|}{Method} & \multicolumn{3}{c|}{PatchCore} & \multicolumn{3}{c}{ReConPatch}    \\
\hline
$\downarrow$ Class\textbackslash{}Aug. Method $\rightarrow$
              & w/o Aug. & w/ Aug.  & Diff     & w/o Aug. & w/ Aug.  & Diff  \\
\hline
Bottle        & 100      & 99.68    & 0.32     & 100      & 100      & 0        \\
Cable         & 99.5     & 96.59    & 2.91     & 99.83    & 99.01    & 0.82     \\
Capsule       & 98.1     & 82.73    & 15.37    & 98.8     & 95.29    & 3.51     \\
Hazelnut      & 100      & 100      & 0        & 100      & 100      & 0        \\
Metal nut     & 100      & 98.92    & 1.08     & 100      & 99.9     & 0.1      \\
Pill          & 96.6     & 92.09    & 4.51     & 97.49    & 94.73    & 2.76     \\
Screw         & 98.1     & 89.75    & 8.35     & 98.52    & 89.96    & 8.56     \\
Toothbrush    & 100      & 95       & 5        & 100      & 99.17    & 0.83     \\
Transistor    & 100      & 98.37    & 1.63     & 100      & 99.33    & 0.67     \\
Zipper        & 99.4     & 95.51    & 3.89     & 99.76    & 99.13    & 0.63     \\
\hline
\multicolumn{1}{c|}{Object classes}
              & 99.17    & 94.86    & 4.31     & 99.44    & 97.65    & 1.79     \\
\hline 
Carpet        & 98.7     & 96.47    & 2.23     & 99.6     & 99.04    & 0.56     \\
Grid          & 98.2     & 87.89    & 10.31    & 100      & 98.5     & 1.5      \\
Leather       & 100      & 100      & 0        & 100      & 100      & 0        \\
Tile          & 98.7     & 96.64    & 2.06     & 99.78    & 100      & -0.22    \\
Wood          & 99.2     & 99.47    & -0.27    & 99.65    & 99.82    & -0.17    \\
\hline
\multicolumn{1}{c|}{Texture classes}
              & 98.96    & 96.09    & 2.87     & 99.81    & 99.47    & 0.34     \\
\hline
\multicolumn{1}{c|}{Average}
              & 99.1     & 95.48    & 3.62     & 99.56    & 98.56    & 1.0      \\
\specialrule{.05em}{0em}{0em} 
\hline
\end{tabular}%
\caption{Anomaly detection performance with data augmentation(random rotate, translate, brightness, contrast and gaussian blur) on the MVTec AD dataset\cite{bergmann2019mvtec}}
\label{table:supple_aug_performance}
\end{table*}

\subsection{Inference time}\label{appendix_subsec:inference_time}
We measured inference time and memory usage of ReConPatch and PatchCore\cite{roth2022towards} with Intel Xeon Gold 6240 CPU and Nvidia GeForce GTX 3090 GPU.
Test images are resized to 256$\times$256 and center cropped to 224$\times$224 on MVTec AD dataset \cite{bergmann2019mvtec}. The target embed dimension was set to 512 in both PatchCore\cite{roth2022towards} and ReConPatch. We experimented for 30 iterations with 10 warm-up times to measure the inference time of PatchCore\cite{roth2022towards} and our proposed ReConPatch. In Table \ref{table:supple_inference_time}, the average inference time of PatchCore\cite{roth2022towards} and ReConPatch are not significantly different.

\begin{table}[hbt!]
\centering
\begin{tabular}{l|c}
\hline
\multicolumn{1}{c|}{Method}
    & Avg. inference time(msec)    \\
\hline
PatchCore                     
    & 37.89    \\
ReConPatch                  
    & 38.09    \\
\hline
\end{tabular}
\caption{Inference time comparison between PatchCore\cite{roth2022towards} and ReConPatch on MVTec AD dataset.}
\label{table:supple_inference_time}
\end{table}

\subsection{Memory efficiency}\label{appendix_sub:memory_usage}
\begin{figure}[h]
\includegraphics[width=1\linewidth]{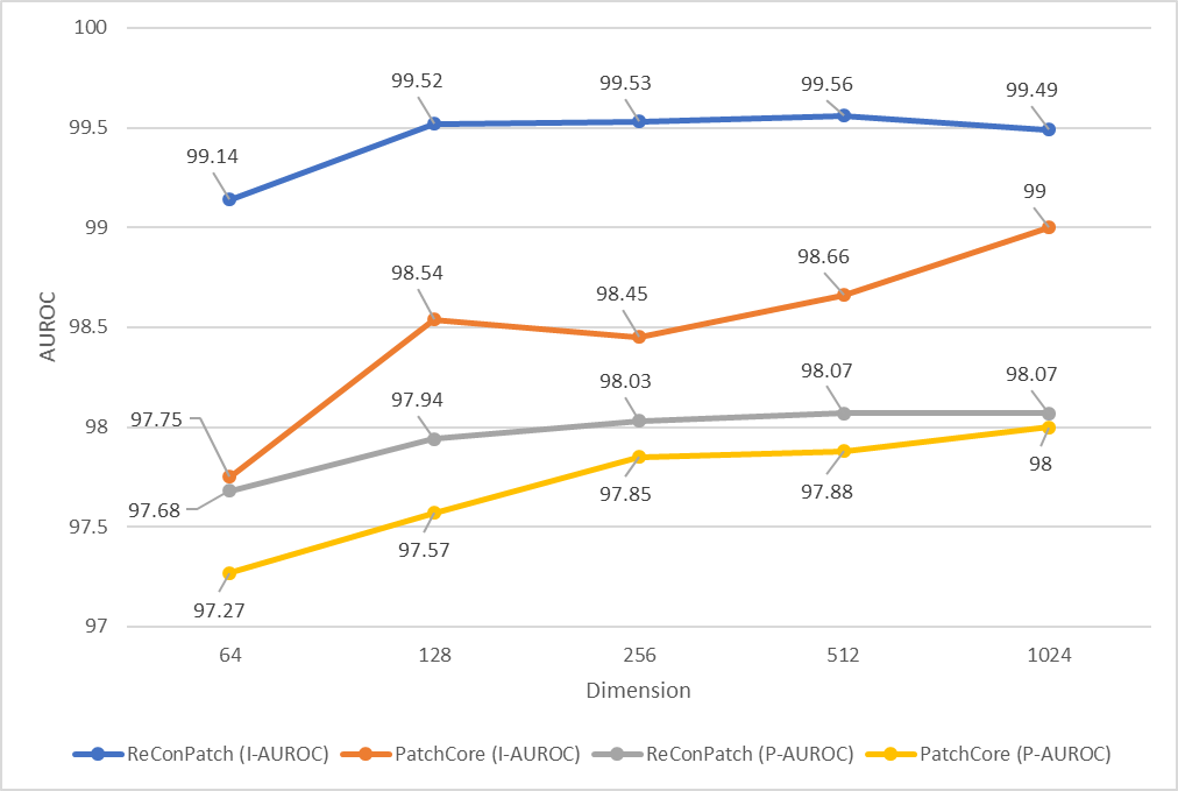}
\caption{The results of Detection AUROC and Segmentation AUROC for the $f$ layer dimension on the MVTec AD dataset\cite{bergmann2019mvtec}. Image size is 224$\times$224, $k=5$, $m=1.0$, $\alpha=1.0$ and coreset subsampling rate is  1\%}
\label{fig:supple_dimension}
\end{figure}

Our proposed ReConPatch algorithm uses a single linear layer which trained with local patch features of normal samples.
In Eq. \ref{eq:supple_memory}, memory usage of memory bank $M_{coreset}$ is affected by dimension of $f$ layer $f_{dim}$ and percentage of coreset subsampling $c_{percentage}$. ${H_{feature}}$ and ${W_{features}}$ are size of outputs in blocks 2 and 3.
The results of anomaly detection and segmentation with various $f$ layer dimension shows in Table \ref{table:dimension_reduction} and Fig. \ref{fig:supple_dimension}.
In Fig. \ref{fig:supple_dimension}, the performance of ReConPatch for anomaly detection (image-level AUROC) and segmentation (pixel-level AUROC) is better than PatchCore\cite{roth2022towards} with same memory bank size.
\begin{equation}\label{eq:supple_memory}
M_{coreset}=f_{dim}*H_{features}*W_{features}*N_{train}*c_{percentage}
\end{equation}

\subsection{Anomaly score maps}\label{appendix_sub:anomaly_score_map}
\begin{figure}[!h]
\centering
\includegraphics[width=0.9\linewidth]{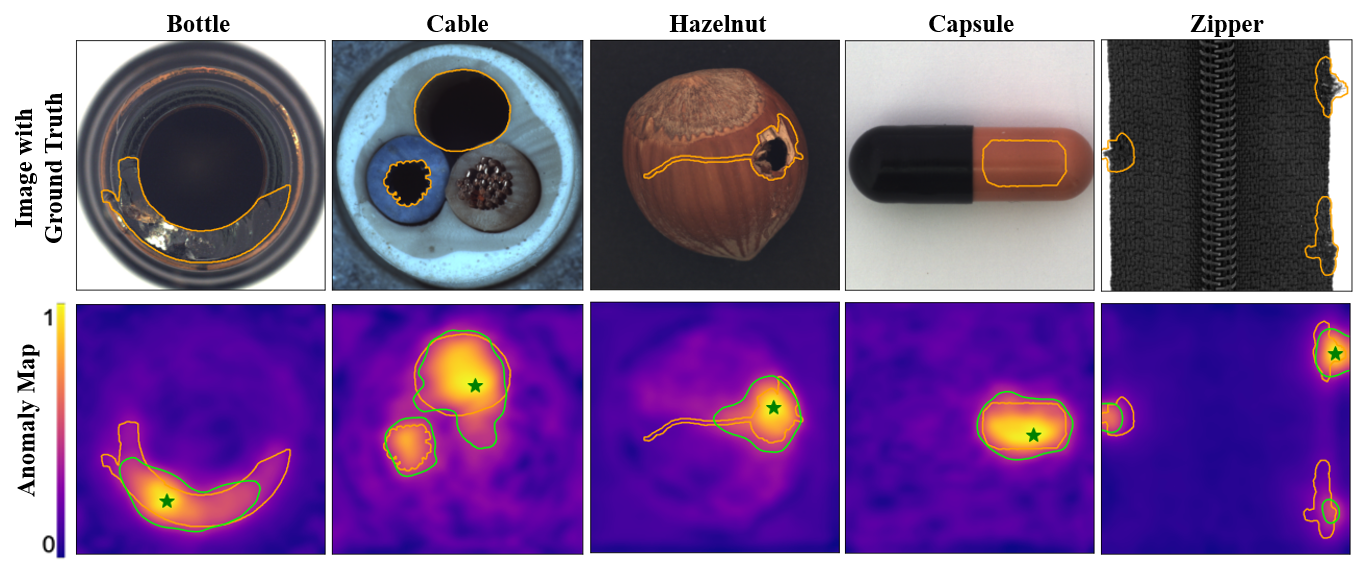}
\includegraphics[width=0.9\linewidth]{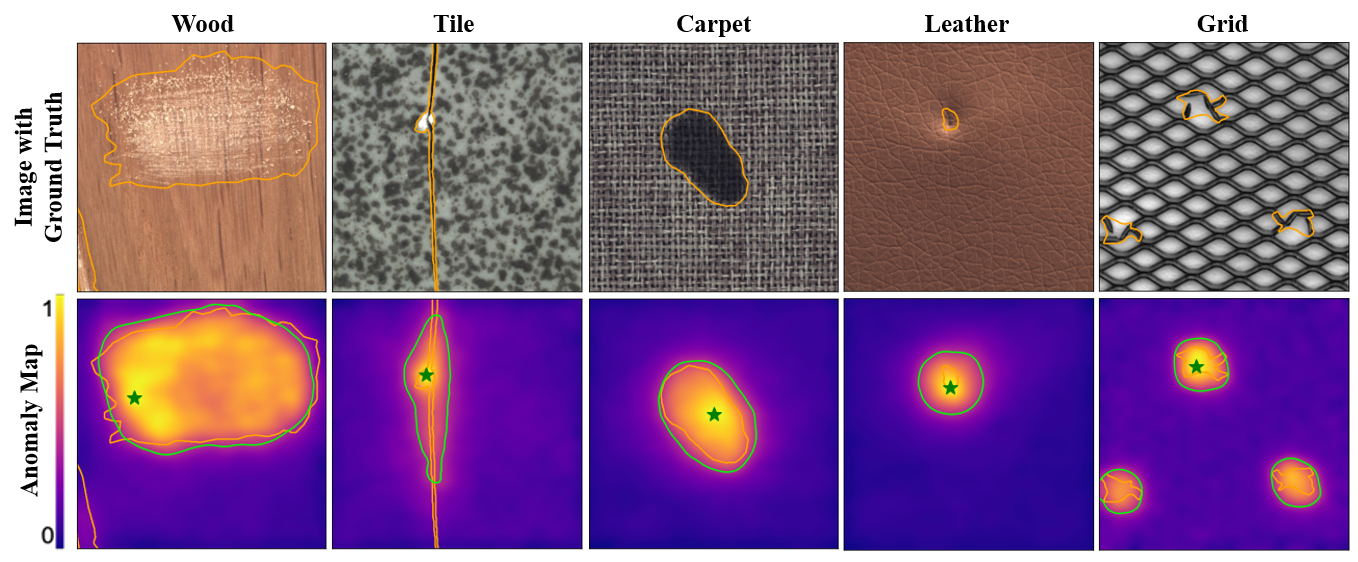}
\caption{Examples of images with anomalies (top) and measured anomaly score maps (bottom) on MVTec AD dataset\cite{bergmann2019mvtec}. The {\color{orange}orange line} depicts the ground truth of the anomalies and the {\color{ForestGreen}green line} depicts thresholds optimizing F1 scores of anomaly segmentation. The {\color{ForestGreen}green star} indicates the maximal location of the anomaly score in the heatmap.}
\label{fig:supple_example_mvtec}
\end{figure}

\begin{figure}[!h]
\centering
\includegraphics[width=0.87\linewidth]{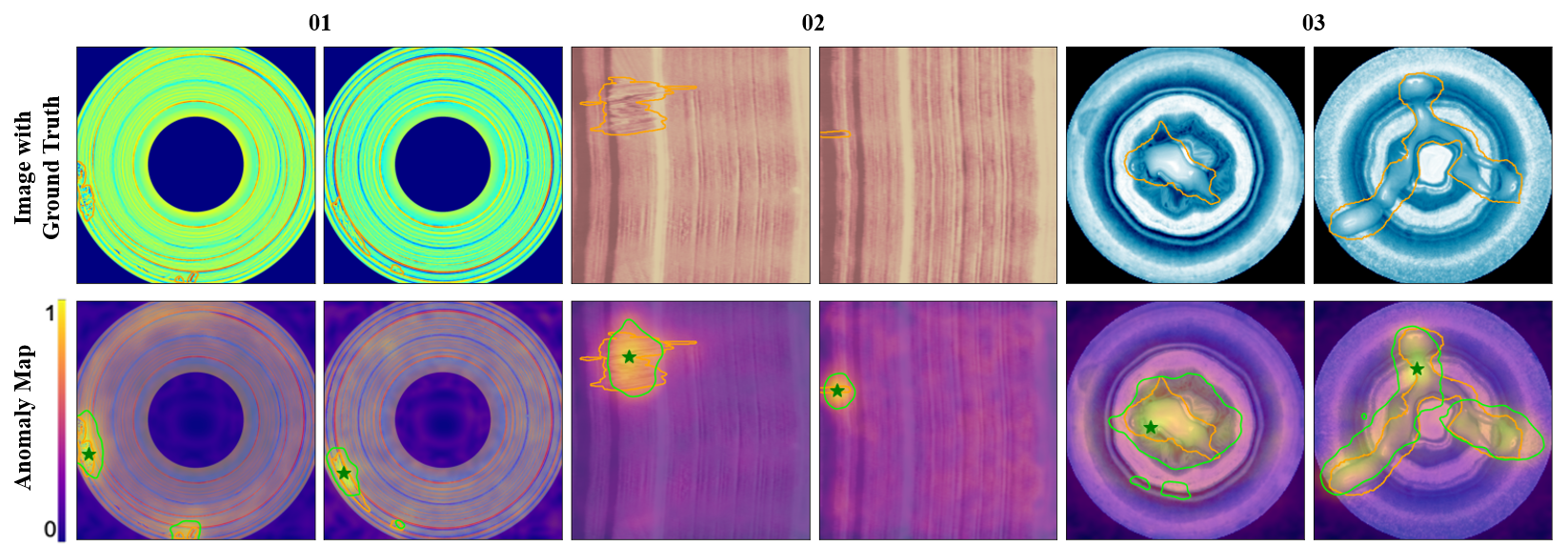}
\caption{Examples of images with anomalies (top) and measured anomaly score maps (bottom) on BTAD dataset\cite{mishra2021vt}. The {\color{orange}orange line} depicts the ground truth of the anomalies and the {\color{ForestGreen}green line} depicts thresholds optimizing F1 scores of anomaly segmentation. The {\color{ForestGreen}green star} indicates the maximal location of the anomaly score in the heatmap.}
\label{fig:supple_example_btad}
\end{figure}

We provide the examples of the anomaly score map of MVTec AD dataset\cite{bergmann2019mvtec} and BTAD\cite{mishra2021vt} dataset along with ground truth (orange line) overlaid input images in Fig.\ref{fig:supple_example_mvtec}, \ref{fig:supple_example_btad}. The anomaly map indicates the regions of the input image where ReConPatch has detected anomalies, with higher values indicating a higher likelihood of an anomaly being present. The {\color{ForestGreen}green line} indicates the threshold which is optimized by the F1 scores of anomaly segmentation, and the {\color{orange}orange line} indicates the ground truth of the anomalies.

\section{Refinement Network for Fine-grained Anomaly Localization}
This section describes the improved performance through the combination of ReConPatch and the refinement network.
As the patch-level score map of ReConPatch is obtained by aggregating features of intermediate layers, it requires the resolution matching between the score map and the input image.
To match the resolution, ReConPatch utilizes the bilinear upsampling and the Gaussian smoothing, but this approach fails to represent abnormal regions with the desired level of detail.
To address this issue, similar to \cite{bae2023pni}, we utilize refinement network to refine anomaly score map of ReConPatch in the MVTec AD dataset. Inspired by \cite{zavrtanik2021draem, yang2023memseg, zhang2023prototypical}, we artificially generate synthetic abnormal images from normal images and DTD dataset \cite{cimpoi2014describing} and utilize focal loss and L1 loss to minimize difference between synthetic anomaly mask and predicted pixel-wise score map for training refinement network with supervised learning.

\subsection{Implementation Details}
Refinement Network contains two residual blocks and several parallel Atrous convolution with different rates (Atrous Spatial Pyramid Pooling \cite{zhang2023destseg}). We train the network using the SGD optimizer for 500 epochs with the batch size of 16, the learning rate of $10^{-2}$, the momentum of 0.9 and the weight decay of $10^{-4}$.
To compare with the ensemble model of ReConPatch, each image was resized and center-cropped to the size described in Section 4.3. 

\subsection{Abnormal Image Simulation}
To train the refinement network, we require the abnormal image and the corresponding mask that indicates the anomaly region. Due to the absence of the abnormal images for the anomaly detection task, we utilize the synthetic abnormal image for the training. In previous work \cite{zavrtanik2021draem, yang2023memseg, zhang2023prototypical}, the synthetic images are generated by mixing the random sampled mask of the normal image with images from the external dataset \cite{cimpoi2014describing}. This abnormal image simulation method works as simulating the textural anomalies. In addition to simulating textural anomalies, MemSeg \cite{yang2023memseg} proposes to simulate the structural anomalies by using the randomly permuted image patches instead of the external image. We compare the performance of refinement network trained on the abnormal image simulation with texture and structure respectively. Table \ref{tab:aug_ablation} shows the average performance over the classes of MVTec AD dataset, which indicates that using the structural anomalies gives the best performance. When analyzing structure and texture from the perspective of anomaly score maps, we observed that the deviation in anomaly scores for structure is smaller than the texture. We evaluated that the smaller deviation cause the better performance for discrimination based on image-level anomaly score.

\begin{table}[!h]
\caption{Evaluating the components of abnormal simulation strategy on the MVTec AD benchmark}
  \centering
  \begin{tabular}{lccc}
    \toprule
    \multicolumn{4}{c}{ReConPatch (WRS-50, 224x224)}                   \\
    \midrule 
    Method & w/o Refinement &\multicolumn{2}{c}{w/ Refinement \cite{zhang2023destseg}}\\
    \midrule 
    Simulation type     & -  & Texture \cite{zavrtanik2021draem}    &  Structure \cite{yang2023memseg}\\
    \midrule 
    Detection & 99.56   & 98.91   & \textbf{99.71} \\
    Segmentation     & 98.18  & 98.43  & \textbf{98.62} \\
    \bottomrule
  \end{tabular}
  \label{tab:aug_ablation}
\end{table}

\subsection{Anomaly detection and localization on MVTec AD}
Based on the previous analysis on the refinement network, we report the performance improvement of refinement network attached to ReConPatch in Table \ref{tab:ensemble}. We only compare the average score over the entire class of MVTec AD dataset, rather than score per each class. Table \ref{tab:ensemble} verifies that the refined anomaly score map achieves better performance in both image-level and pixel-level AUROC.

\begin{table}[!h]
\centering
\begin{tabular}{@{}lcccc@{}}
\toprule
Ensemble Backbone & \multicolumn{4}{c}{WRN-101 \&   RNext-101 \& DenseN-201}                \\ \midrule
Image size        & \multicolumn{2}{c}{480 × 480}               & \multicolumn{2}{c}{320 × 320}               \\ \midrule
Method            & ReConPatch & ReConPatch(w/ refine) & ReConPatch & ReConPatch(w/ refine) \\ \midrule
Detection         & 99.72      & \textbf{99.86}                 & 99.67      & \textbf{99.78}                 \\
Segmentation      & 98.67      & \textbf{99.20}                 & 98.36      & \textbf{98.96}                 \\ \bottomrule
\end{tabular}
\caption{Comparison of ensemble model anomaly detection(image-level AUROC) and segmentation (pixel-level AUROC) performance on the MVTec AD dataset}
\label{tab:ensemble}
\end{table}

\begin{figure}[!h]
  \centering
  \includegraphics[width=\columnwidth]{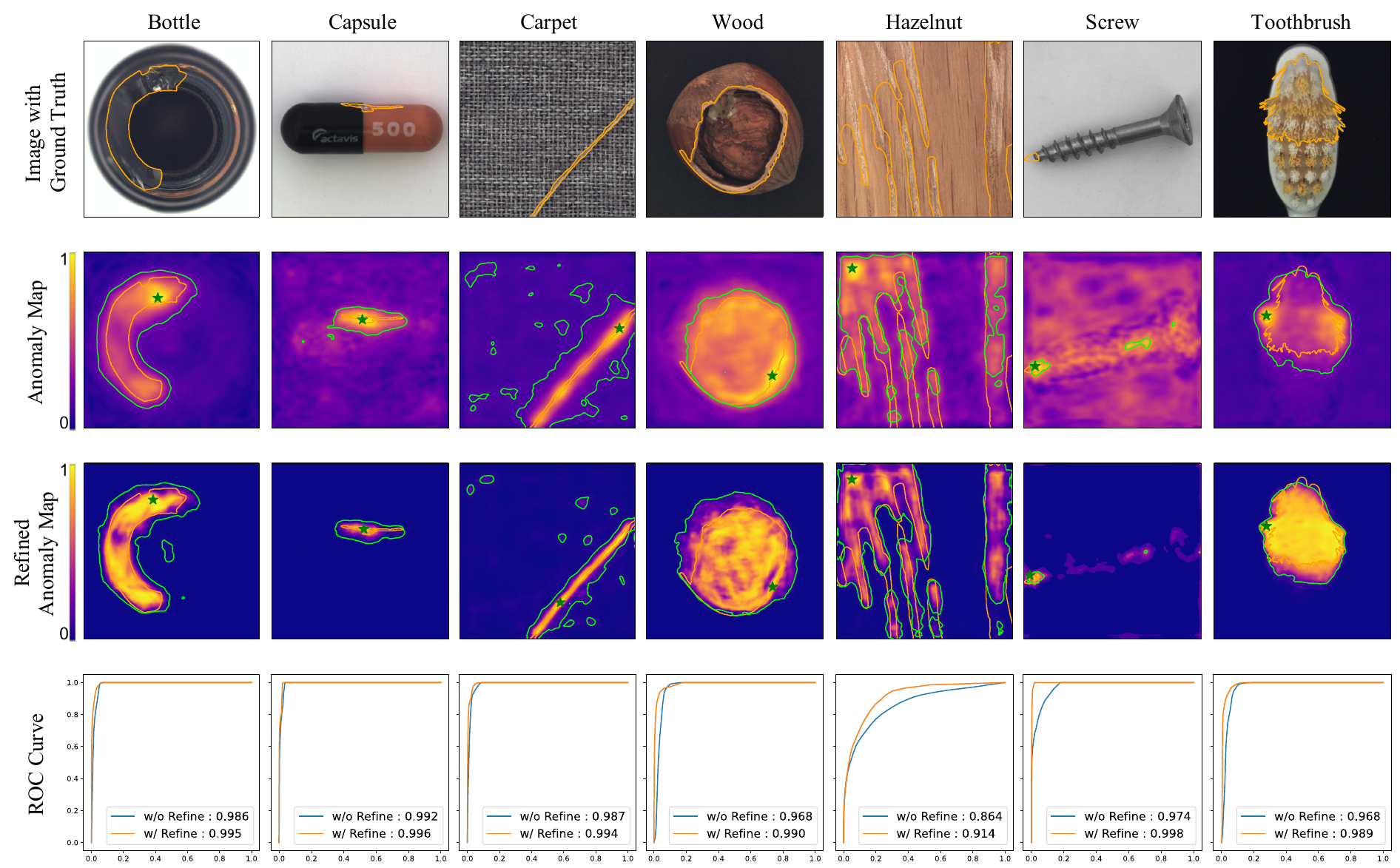}
  \caption{Visualization of the anomaly score map comparing before and after the refinement.}
  \label{fig:refine_visualize}
\end{figure}

We also visualize the qualitative difference in the anomaly score map in Figure \ref{fig:refine_visualize}. The second row and the third row depict the score map of ReConPatch without and with refinement network respectively. Refined score maps show more fine-grained localization of defects in the images. The performance improvement also rises from reducing the anomaly scores on the irrelevant locations, such as background. The bottom row shows the pixel-wise roc curve for each anomaly score map, which indicates that the refined score map surpasses the score map without refinement.

\end{document}